\documentclass{article}
\usepackage{arxiv}
\usepackage{algorithm}
\usepackage{algpseudocode}
\usepackage[style=numeric]{biblatex}
\bibliography{references,other_methods}
\usepackage{subfigure}
\usepackage{graphicx} 

\begin{document}
%
\title{Federated K-Means clustering}

\author{Swier Garst \\
	Delft Bioinformatics Lab\\
	Delft Univserity of Technology\\
	Delft, the Netherlands \\
	\texttt{s.j.f.garst@tudelft.nl} \\
	\And
	Marcel Reinders \\
	Delft Bioinformatics Lab\\
	Delft University of Technology\\
	Delft, the Netherlands \\
}
\date{}
\maketitle              

\begin{abstract}

Federated learning is a technique that enables the use of distributed datasets for machine learning purposes without requiring data to be pooled, thereby better preserving privacy and ownership of the data. While supervised FL research has grown substantially over the last years, unsupervised FL methods remain scarce. This work introduces an algorithm which implements K-means clustering in a federated manner, addressing the challenges of varying number of clusters between centers, as well as convergence on less separable datasets.
\keywords{Federated Learning  \and K-Means clustering \and Distributed machine learning.}
\end{abstract}

\section{Introduction}
Nowadays, lots of data is being generated in a distributed fashion. Mobile phones and other personal devices such as smart watches enable the collection of massive amounts of data. If made accessible, this data could prove useful for improving the performance of the services provided by these devices. However, due to a growing concern on data privacy, more and more users of these devices are hesitant in sharing their data. Furthermore, regulations such as the General Data Protection and Regulation (GDPR) act prevent the collection of data of this kind in bulk. Federated learning (FL) (\cite{fedAvg}) was introduced as a solution to this problem. In short, instead of pooling data to train a single model, instances of a model are being shared to data owners (clients), which then train the model on their local data. Then, these trained models are sent back to the central server, which aggregates the results. Next, a new round begins with the server sending out the updated models. This cycle continues until convergence.
\\\\
Over the past couple of years, research has shown FL to be a promising technique, reaching performances comparable to a central approach in which all data of the clients is pooled at a single location\cite{FLbench1} \cite{FLbench2}. The vast majority of the federated learning research has been focusing on the supervised learning paradigm. Little work has been done on unsupervised federated learning methods, even less so when specifically looking into clustering techniques \cite{FLDirections} \cite{FLHype}. One of these clustering techniques is k-means clustering \cite{KMeans}. In a federated learning setting, k-means clustering can be described as trying to find overarching cluster means over data which is distributed among different datasets (clients). Only a few papers discuss possible options for federated k-means clustering. Kumar et al. introduce a method which assumes availability of some data at the central server to pretrain the model, which is not always feasible\cite{Kumar2020}. Hou et. al. use secure multiparty computation and blockchain to share encrypted data on which k-means is being performed \cite{Hou2021}. However, for some cases (e.g. in the medical domain), even sharing encrypted data might not be acceptable. 
\\\\
Bringing k-means into the federated domain comes with a specific challenge: not all clients need to have the same amount of clusters in their data. Therefore, they might also not have data from each cluster. Due to this possible heterogeneity, a way of matching clusters from different clients is required. Liu et. al. introduced a method for federated k-means in a setting where each client only holds one sample \cite{Liu2020}. Although this is highly valuable in the cross device setting, it does not address the challenges of varying amounts of local clusters. Servetnyk et. al. propose a dual averaging approach for k-means clustering, which does not seem to address the challenge on how to match clusters from different clients \cite{FLKMeans}. Dennis et. al. propose a one-shot method by doing a k-means clustering over the cluster means found locally \cite{Dennis2021}. However, their experiments assume an equal amount of local clusters in the data available, as well as highly separable clusters. Furthermore, the number of local clusters is an input parameter to their method, even though it is not always possible to determine this. Nevertheless, their choice to run a clustering centrally over cluster means from the local datasets inspired us to do this in an iterative approach. Next to that, we introduce a method that deals with a variable amount of local clusters. We show that both changes lead to improvements, especially in settings where the amount of local clusters is highly varying between clients, or when clusters are less separable. Moreover, we show that in a 2-dimensional setting, a clustering of similar quality can be obtained compared to a conventional k-means clustering (on a pooled dataset). 


%
\section{Methods}

\begin{table}[h]
    \centering
    \caption{notations used}
    \label{tab:notation}
    \begin{tabular}{c|c}
        symbol &  description \\
    \hline
        $X_i$ & data on client $i$ \\
        $K_g$ & global number of clusters \\
        $K_i$ & number of clusters on client $i$ \\
        $C_g$ & global cluster means \\
        $C_i$ & cluster means of client $i$ \\
        $M$ & total amount (sum) of local clusters \\
        $S_i$ & amount of samples for each cluster on client $i$ \\
        $N$ & total amount of clients
    \end{tabular}
\end{table}

\begin{algorithm}
\caption{The federated kmeans algorithm \\ 
\textbf{Input: $K_g$} }
\label{alg:fkmeans}
\begin{algorithmic}[1]

\State \textbf{Init:}
\State  \textbf{    }\textbf{on each client $i \in N$ do:}
\State  \textbf{    }\textbf{    }$K_i$ = $K_g$
\State  \textbf{    }\textbf{    }$S_i$, $C_i$ = kmeans++\_init($X_i$, $K_i$) \Comment{get cluster means using kmeans++ initialization}
\State  \textbf{    }\textbf{    }send $S_i$, $C_i$ to server

\State \textbf{For each round $r$ do:}
\State \textbf{    }\textbf{On server do:}
\State \textbf{    }\textbf{    }$C_l$ = [$C_1 | C_2 | .. | C_M$] \Comment{Concatenate all local cluster means}
\State \textbf{    }\textbf{    }$S_l$ = [$S_1 | S_2 | .. | S_M$]\Comment{Repeat for sample amounts per cluster}
\State \textbf{    }\textbf{    }$C_g$ = kmeans($C_l$, $K_g$, weights = $S_l$) \Comment{Obtain new global clusters using kmeans}
\State \textbf{    }\textbf{    }send $C_g$ to all clients

\State \textbf{    }\textbf{On each client $i \in N$ do:}
\State \textbf{    }\textbf{    }$C_i$ = $C_g$
\State \textbf{    }\textbf{    }$S_i$ = kmeans\_assign($X_i$,$C_i$) \Comment{Determine empty clusters}
\State \textbf{    }\textbf{    }$C_i$ = $C_i$[s != 0 for s in $S_i$] \Comment{Drop empty global clusters}
\State \textbf{  }  $K_i$ = size($C_i$)
\State \textbf{    }\textbf{    }$S_i$, $C_i$ = kmeans($X_i$, $K_i$, init = $C_i$) \Comment{run kmeans from remaining global cluster means}
\State \textbf{    }\textbf{    }send $S_i$, $C_i$ to server
\end{algorithmic}
\end{algorithm}




\noindent Notation used throughout this section is found in table \ref{tab:notation}. The pseudocode for our proposed federated k-means algorithm (FKM) can be found in algorithm \ref{alg:fkmeans}. The algorithm can be divided into two parts: an initialization step, in which we generate initial cluster means on each client using k-means++ initialization \cite{KMeans++}, and an iterative k-means step in which clients communicate their cluster means to the server, which aggregates these means into a 'global' set of means, which then gets redistributed to the clients for the next k-means iteration. See appendix \ref{sec:background} for background on k-means and k-means++.
\\\\
\textbf{Determining the amount of local clusters.} \\
\noindent While the global amount of clusters is set (main parameter $k$ of the k-means procedure), it is not a given that each client has data for each of these clusters. In other words, the number of clusters between clients can differ, and is not necessarily equal to the number of clusters in the pooled data. In order to solve this problem, we determine which global clusters correspond to local data in each round on each client. Before a client applies a new k-means step locally, it assigns its data to the global cluster means it has received (line 14). Next, clients check if there are empty clusters, i.e. cluster means which did not get any points assigned to them. If so, clients discard these empty clusters (line 15). The remaining (global) cluster means are then used as initialization for the next local k-means step (line 16). This way, k can locally become smaller when running k-means on the clients.
\\\\
\textbf{Cluster alignment} \\
\noindent After each client has calculated one iteration of k-means (not until convergence, to avoid local minima) on their local data (each with their own amount of local clusters), they send their cluster means as well as the amount of samples per cluster back to the server. The server then concatenates all cluster means, and aggregates them. It does so by running a k-means clustering on the received local means until convergence (using the global k parameter), to align clusters from different clients to each other. This global k-means is weighted by the amount of samples per cluster found, such that a cluster with lots of samples in it will have a bigger impact on the aggregation step compared to a cluster with fewer samples. That is, we modify the k-means objective function (see appendix \ref{sec:background} for the original) into:
\begin{equation}
    F_{km} = \sum_{j=0}^M \min_{C_{i} \in C_g}(S_j||C_j - C_{i}||^2)
\end{equation}
where $S_j$ is the amount of samples corresponding to local cluster $C_j$. Note that the \textit{cluster means} sent back by the clients are at the server used as the \textit{samples} for clustering using k-means. Doing the aggregation with a k-means clustering, we solve the cluster alignment problem, since similar clusters will be close to each other and thus merged by the global k-means step.
\\\\
Because the amount of samples have to be reported to the central server, there exists a privacy risk if a client finds a cluster with only one sample in it. To prevent this, any clusters holding less than $p$ samples (we used $p = 2$ throughout this work) are simply omitted from the list of means sent to the server.


\section{Results}

\begin{figure}[!htb]
\centering
    \subfigure[]{
        \includegraphics[width=0.47\textwidth]{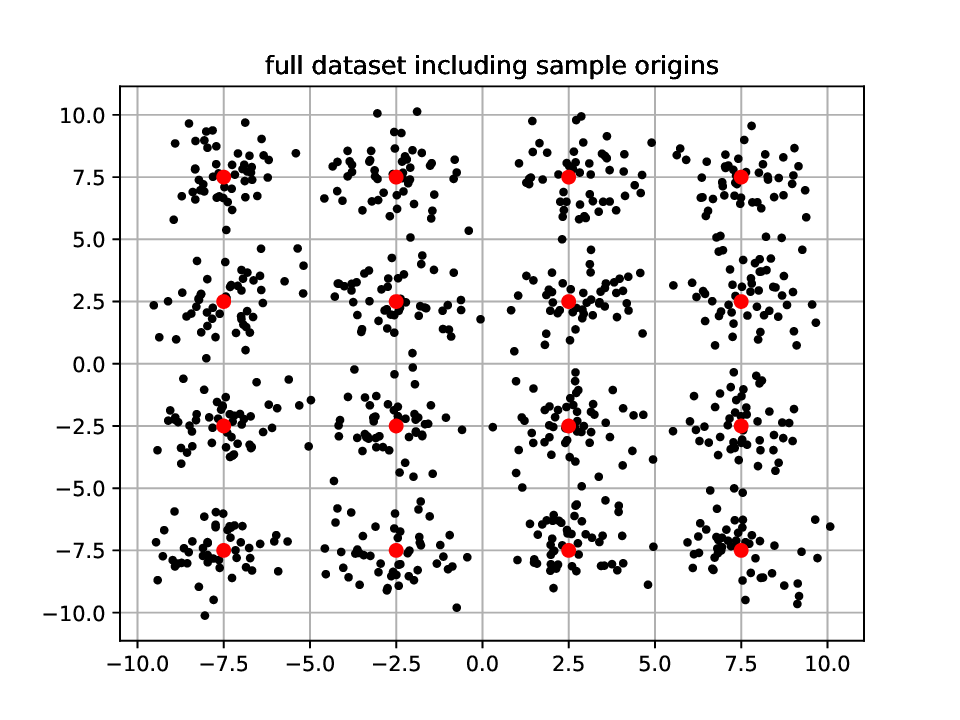}
        \label{fig:full_dset}
    }
        \subfigure[]{
        \includegraphics[width=0.47\textwidth]{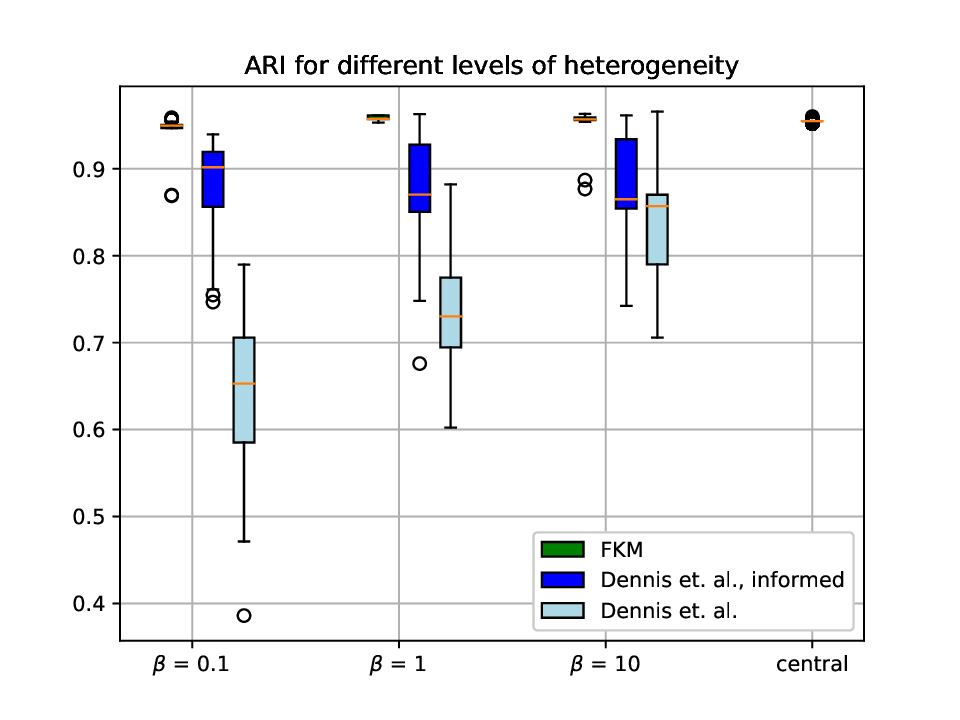}
        \label{fig:synth_orig}
    }
    \subfigure[]{
        \includegraphics[width=0.3\textwidth]{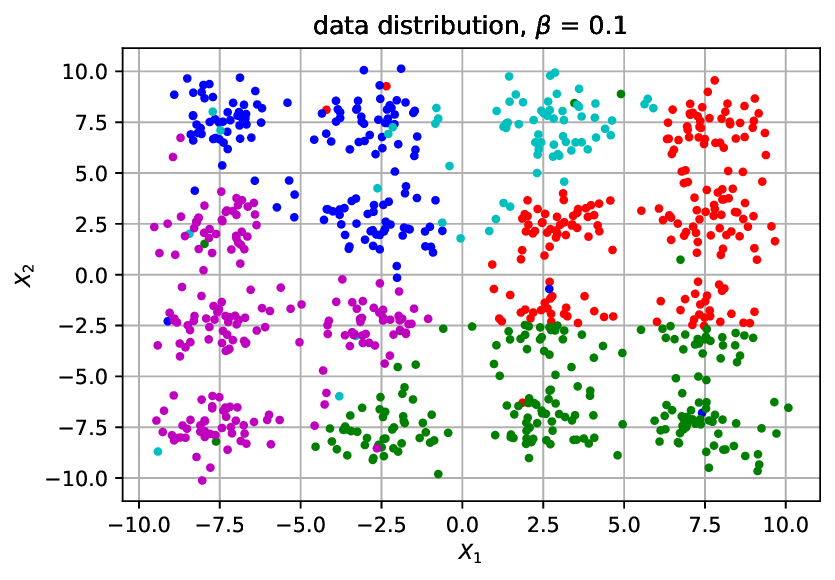}
        \label{fig:dset_dist0.1}
    }
    \subfigure[]{
        \includegraphics[width=0.3\textwidth]{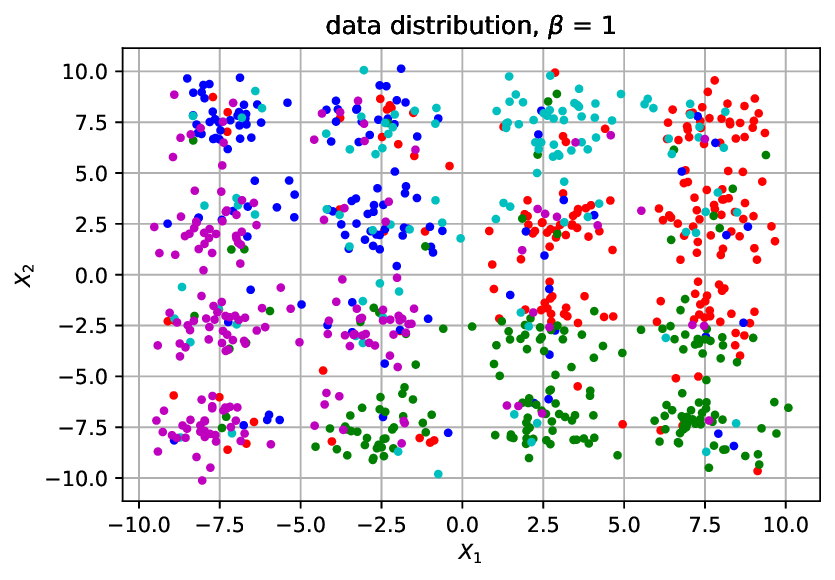}
        \label{fig:dset_dist1}    
    }
    \subfigure[]{
        \includegraphics[width=0.3\textwidth]{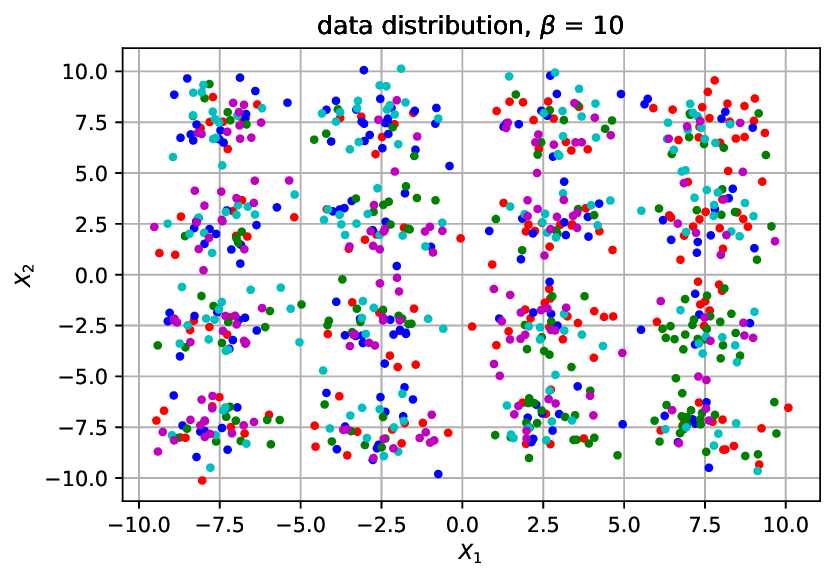}
        \label{fig:dset_dist10}
    }
    \vfill
1
    
    \caption{The regular synthetic datasets. (a) shows the original sampling of the regular synthetic dataset, with the defined cluster means (from which the data are generated using a normal distribution N(0,1)) in red. (b) shows ARI results on all three datasets. (c) until (e) shows how this dataset is distributed over five different clients using different values of $\beta$. Different colors indicate the different clients. }
    \hfill
    \label{fig:dset}
\end{figure}

We compared our federated k-means (FKM) with a k-means clustering that is executed on all data centrally, as well as to one-shot the method of Dennis et. al \cite{Dennis2021}. Our first set of experiments is on simulated data, such that ground truth labels of the cluster centers is known. We therefore calculate the Adjusted Rand Index (ARI) for both central and federated approaches with respect to the labelled samples data. Since there are no labels for the clustering in the FEMNIST experiment, the silhouette score was used instead. In some cases, we added an "informed" setting for Dennis et. al., in which we set $K_{l}$ such as to achieve the highest ARI score by exhaustive search. In all other cases, we run their method using $K_{l}$ = $K_{g}$, as the ARI score is only available when ground truth labels are known, which is not always the case.
\\\\
\noindent \textbf{Clients holding different parts of the data.}\\
\noindent In order to validate the FKM algorithm, a synthetic two-dimensional dataset was generated. The generation procedure is taken from Servetnyk et. al. \cite{FLKMeans}. Sixteen cluster centers were chosen with an equal distance (here 5) from one another, see fig. \ref{fig:full_dset}. Then, 50 data points were sampled around each cluster center using a normal distribution (with variance 1). This data was then distributed among four clients in the following way: First, each client is assigned a 'location' within the field ($X_1,X_2 \in (-12.5,12.5)$). From there, the probability $P$ that a data point would be assigned to a certain client scales inversely with the euclidean distance $d$ to that datapoint:

\begin{equation}
    P = 1 - exp(-\frac{\beta}{d})
\end{equation}

\noindent where $\beta$ is a parameter which can be tuned to promote more or less heterogeneity in the data separation. Differing from \cite{FLKMeans}, if a data point happens to be assigned to multiple clients, it instead gets assigned at random. 
\\\\

\noindent We wanted to explore the influence of data heterogeneity, i.e. a varying amount of clusters per client. To do so, we generated three versions of this dataset, with $\beta = 0.1, 1, 10$. See figure \ref{fig:dset} c-e for the final distributions. Note that $\beta$ only changes which points get assigned to which client, meaning that it does not influence the performance for the central case. Figure \ref{fig:synth_orig} shows that our method is able to attain performance similar to a centralized k-means clustering, while outperforming Dennis et. al., regardless of tuning of the $K_l$ parameter. Performance of our FKM approach seems to be independent of $\beta$ (in constrast to the method of Dennis et. al.), meaning that our algorithm is robust to having varying cluster amounts between clients.

\begin{figure}[!htb]
    \centering
    \subfigure[]{
    \includegraphics[width=0.3\textwidth]{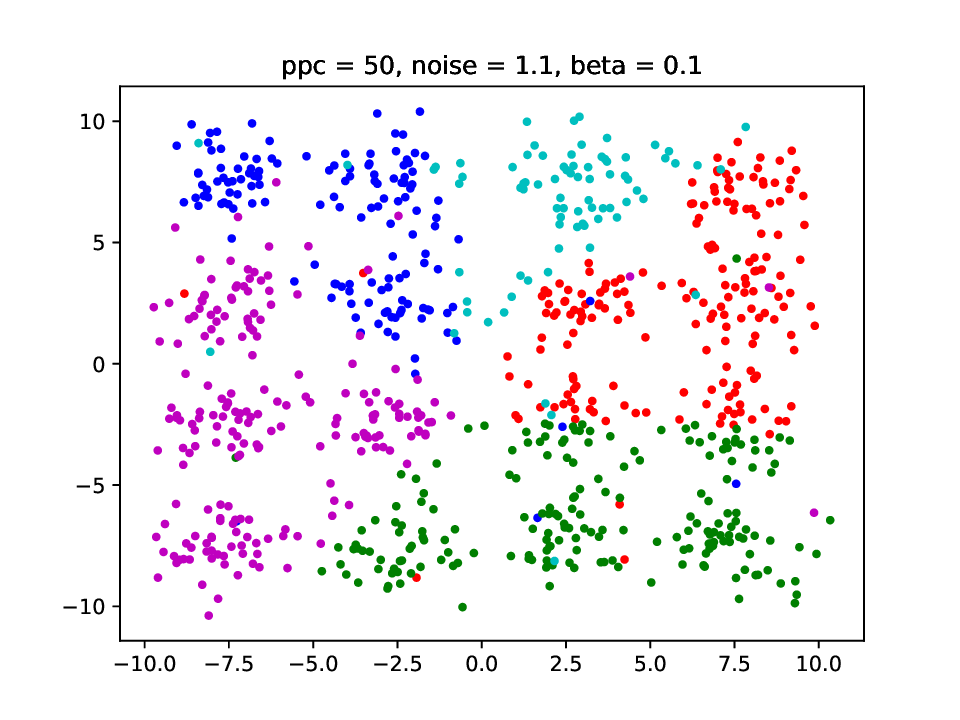}
    }
    \subfigure[]{
    \includegraphics[width=0.3\textwidth]{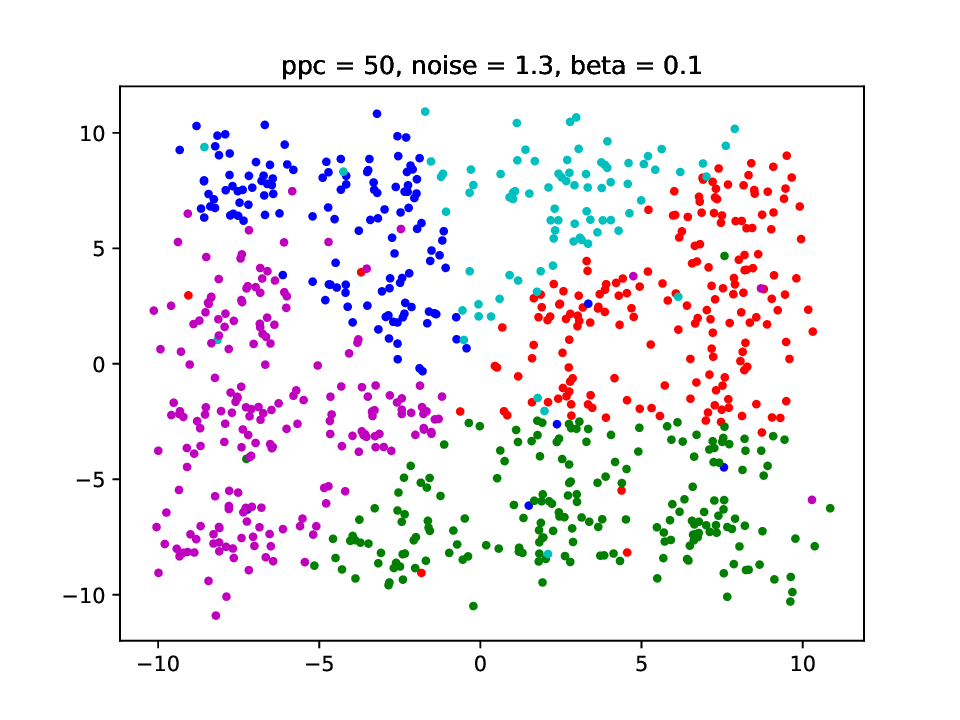}
    }
    \subfigure[]{
    \includegraphics[width=0.3\textwidth]{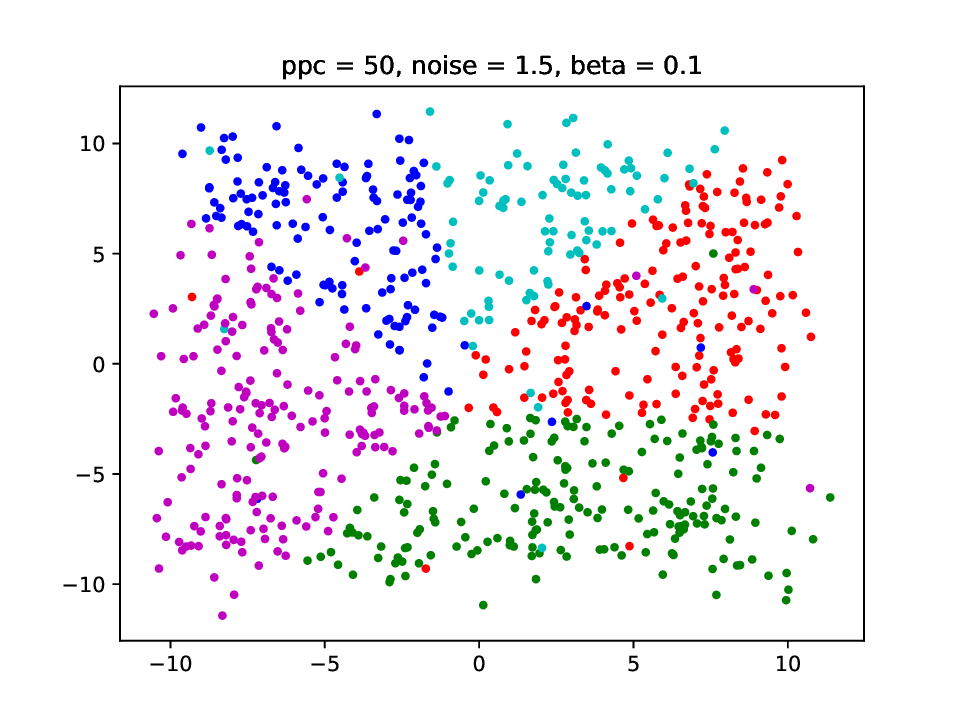}
    }
    \subfigure[]{
    \includegraphics[width=0.3\textwidth]{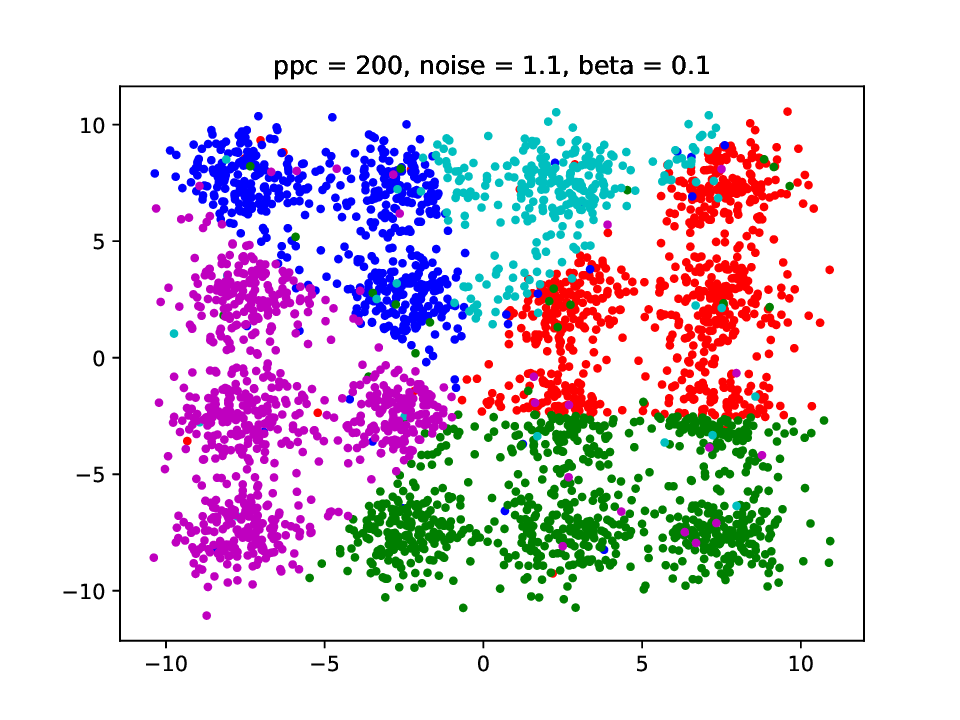}
    }
    \subfigure[]{
    \includegraphics[width=0.3\textwidth]{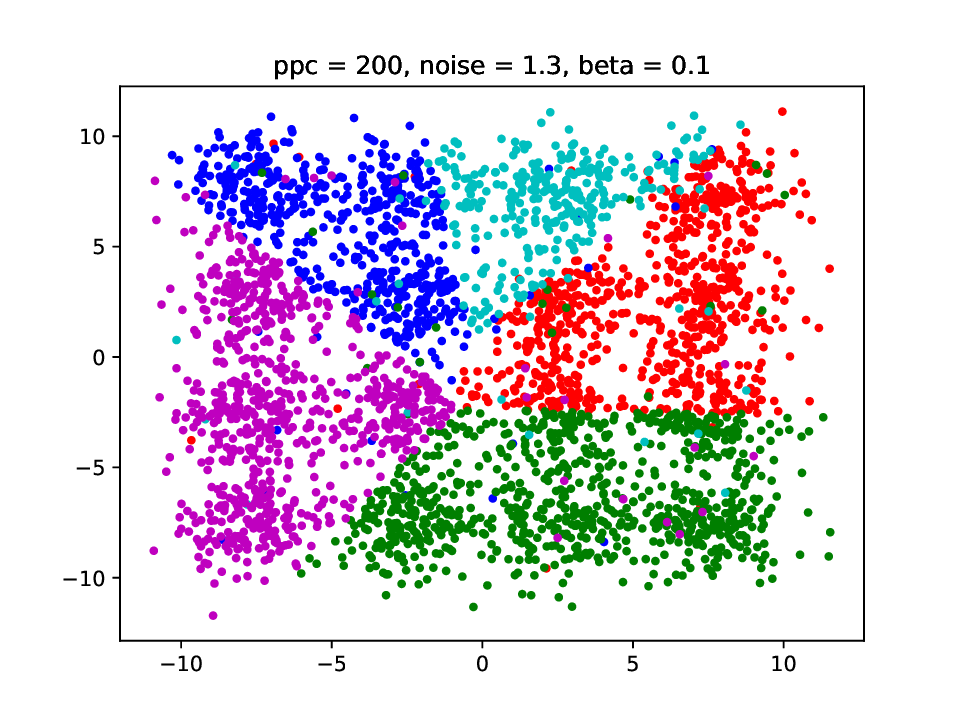}
    }
    \subfigure[]{
    \includegraphics[width=0.3\textwidth]{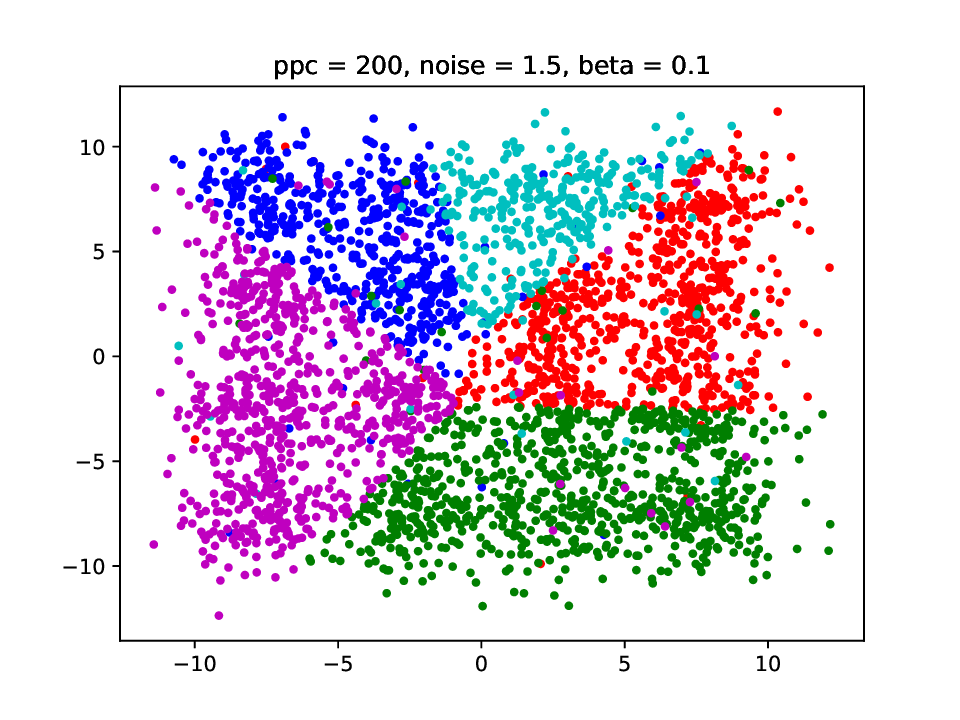}
    }
    \caption{Some of the data distributions of the simulated datasets with increasing levels of noise (columns), using 50 or 200 points per cluster (rows).}
    \label{fig:abl_dist}
\end{figure}

\begin{figure}[!htb]
    \centering
    \subfigure[]{
        \includegraphics[width=0.3\textwidth]{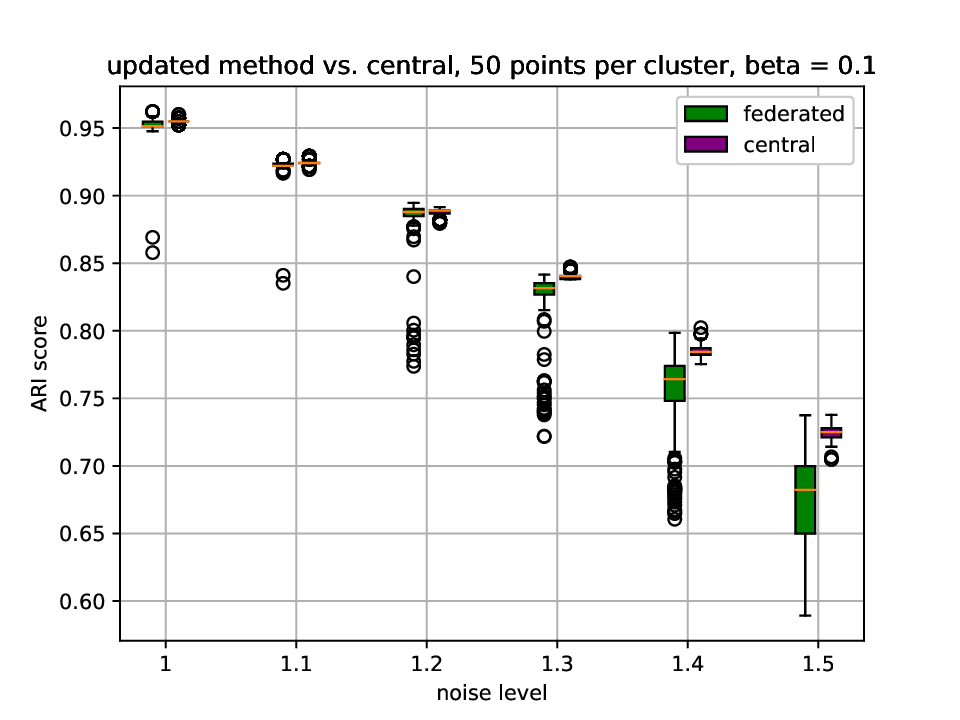}
        }
    \subfigure[]{
        \includegraphics[width=0.3\textwidth]{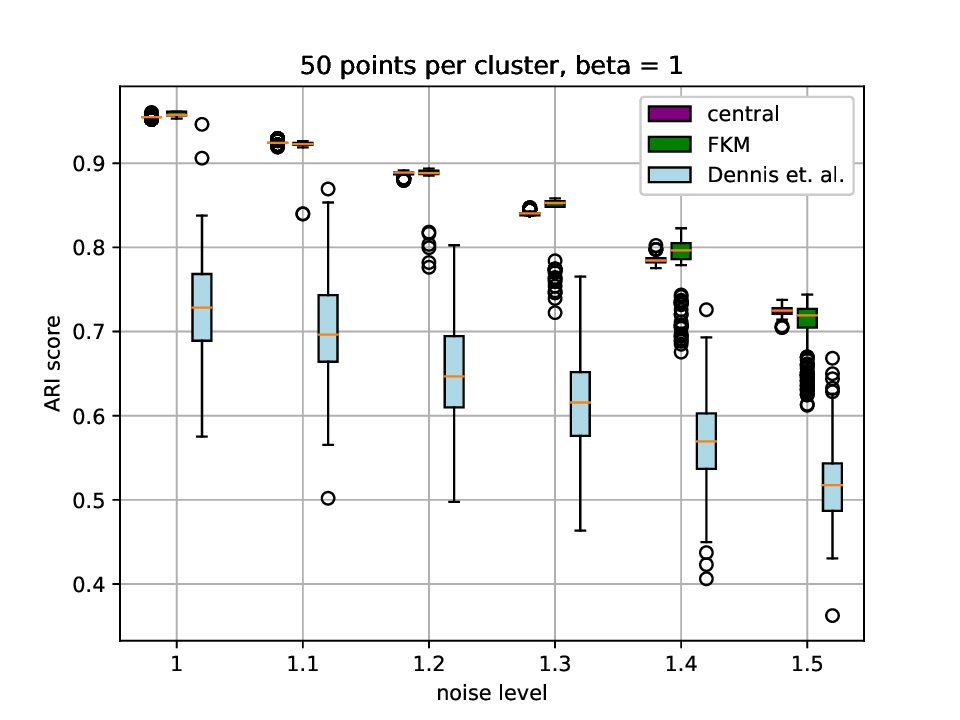}
    }
    \subfigure[]{
        \includegraphics[width=0.3\textwidth]{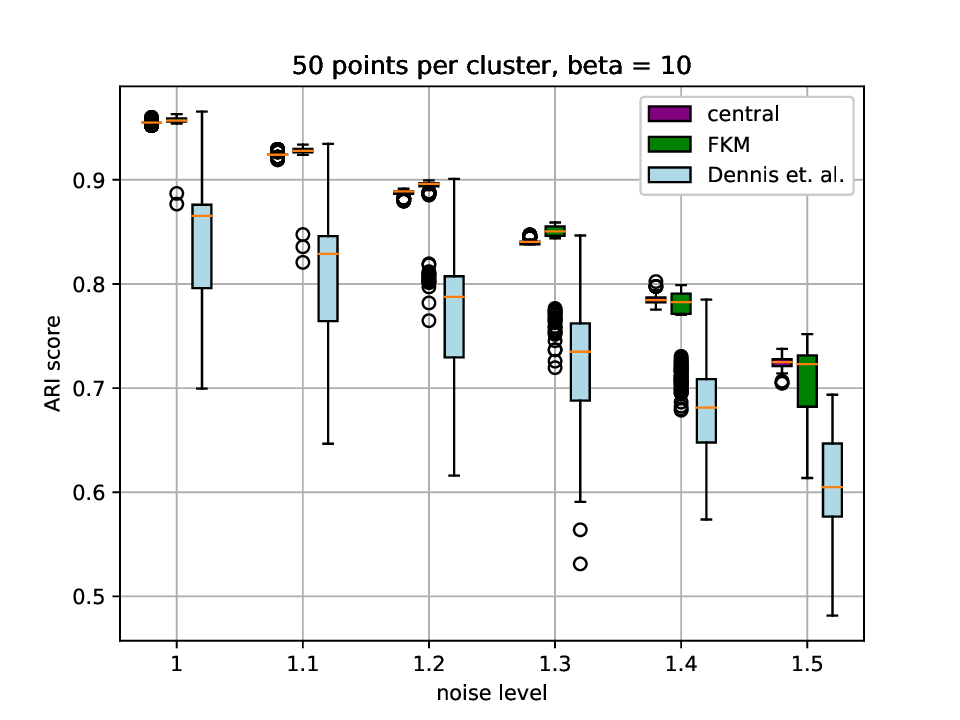}
    }    
    \subfigure[]{
        \includegraphics[width = 0.3\textwidth]{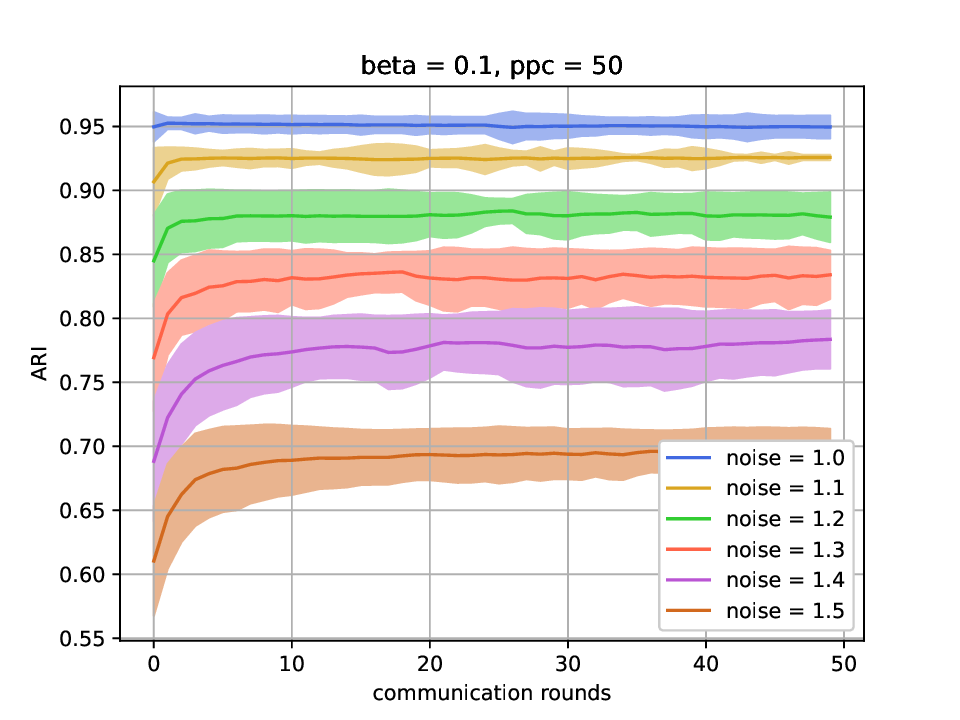}
    \label{fig:conv1}
    }
    \subfigure[]{
        \includegraphics[width = 0.3\textwidth]{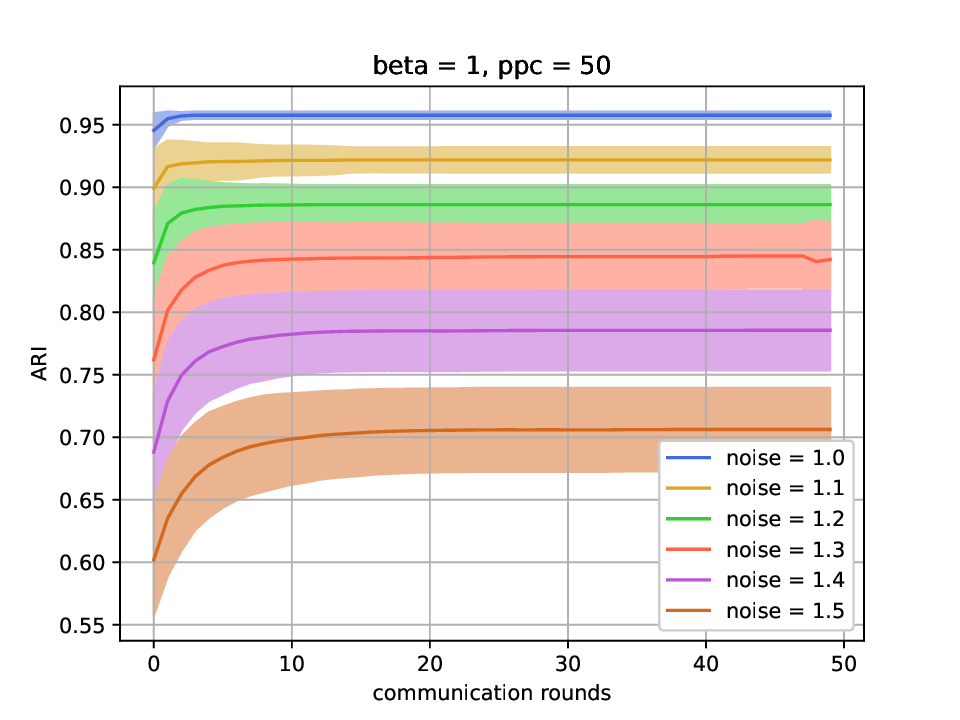}
    }
    \subfigure[]{
        \includegraphics[width = 0.3\textwidth]{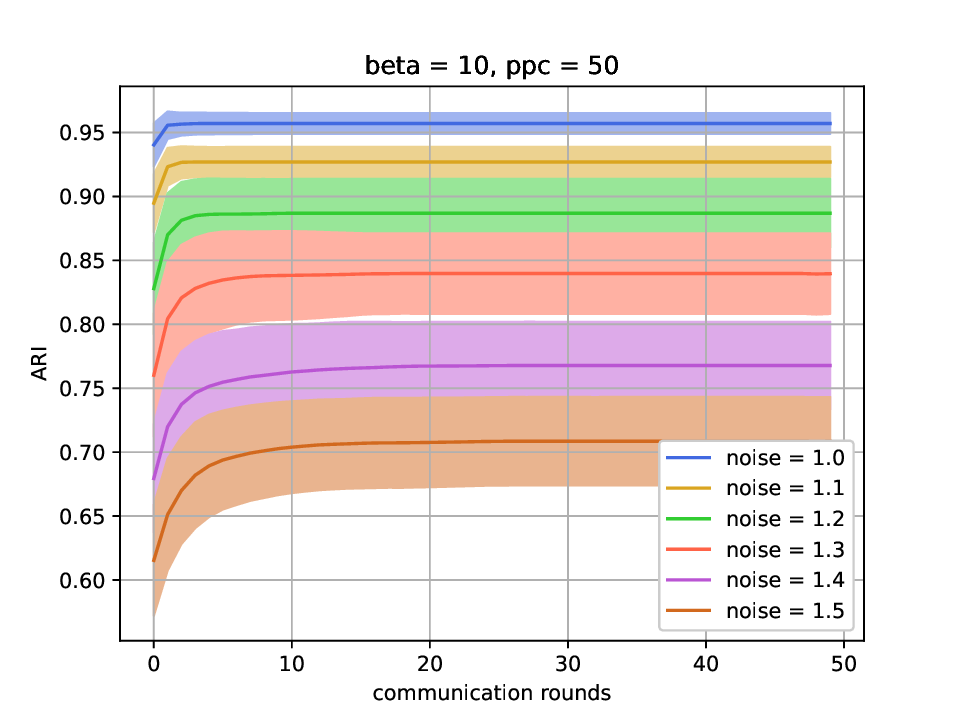}
        \label{fig:conv3}
    }
    \caption{Clustering results on the synthetic dataset when using different levels of noise for different values of $\beta$. (a) to (c) show the final ARI scores for beta = 0.1, 1 and 10, respectively. (d) to (f) show how the ARI score for FKM converges over time, each corresponding to the figure above it.}\label{fig:abl}
\end{figure}

\clearpage
\noindent \textbf{Increasing levels of noise.} \\
Next, we explored the effect of having noisier clusters. We recreated the regular synthetic dataset, but varied the standard deviation from which samples are being generated, from 1 to 1.5 (original used 1). Figure \ref{fig:abl_dist} shows the effect. We generated these datasets twice, once with 50 points per cluster and once with 200 points per cluster. 
\\\\
Results on these datasets are shown in figure \ref{fig:abl}. For both central and federated clustering, the ARI scores go down for higher noise levels. This is expected, as there will be more points ending up closer to the cluster they did not originally belong to, meaning that even if kmeans finds the original cluster means perfectly, the label assignment will be off. Therefore, the relative difference between federated and central clustering is more important than the absolute ARI scores. Our method attains a similar average performance; however, variance seems to increase compared to centralized clustering. Furthermore, for $\beta = 0.1$, mean ARI decreases compared to central clustering at high noise levels, meaning that a setting with high noise as well as high cluster variability is still a hard challenge for our federated k-means algorithm.
\\\\
Regardless, performance does seem to increase significantly as compared to the method of Dennis et. al. . This can partly be due to our ability to iterate. Figure \ref{fig:conv1} to \ref{fig:conv3} shows that, especially for noisier datasets, there is a large benefit in being able to iterate more often. The amount of points per cluster does not seem to influence ARI score significantly, see appendix \ref{sec:exresults}.

\begin{figure}[!htb]
\centering
    \subfigure[]{
        \includegraphics[width=0.47\textwidth]{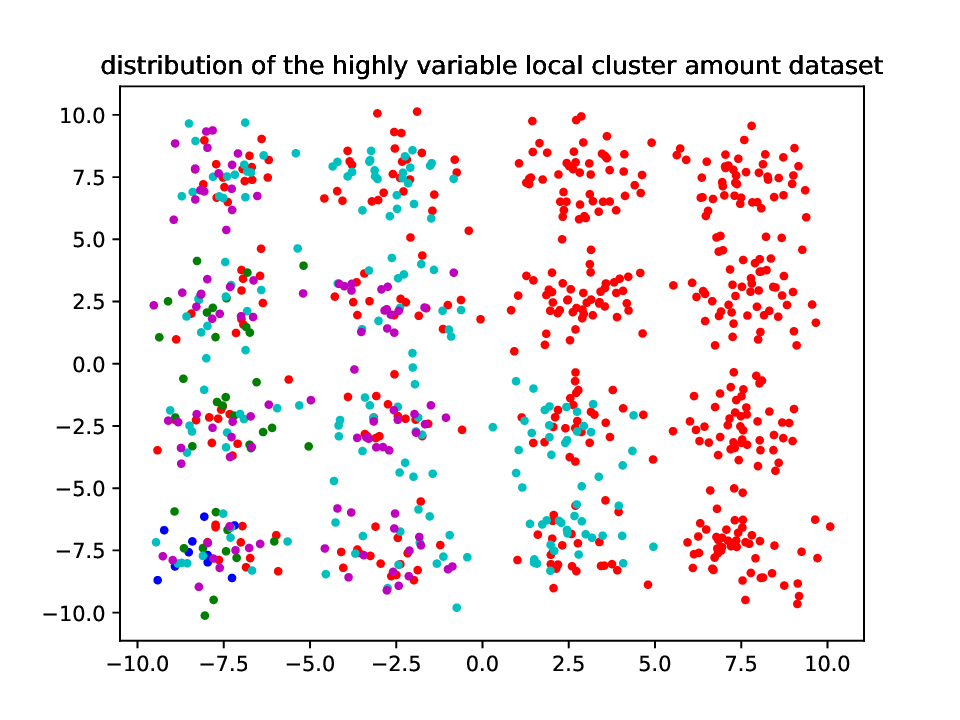}
        \label{fig:ci_dist}
        }
    \subfigure[]{
        \includegraphics[width=0.47\textwidth]{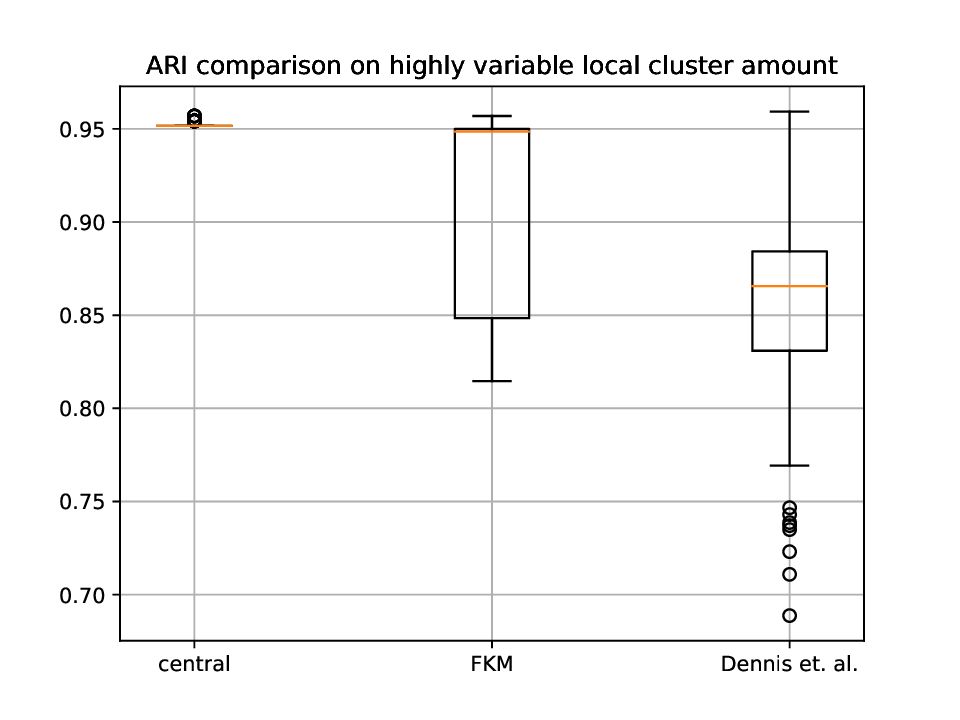}
        \label{fig:ci_results}   
    }
    \caption{Assessment of the method on data with a large variability of local clusters per client. (a) shows the distribution per client, (b) the ARI results for different methods.}
\end{figure}

\noindent \textbf{High variability in number of local clusters.}\\
Next we wanted to explore the effect of having an even more variable local $k$. We used the same data as generated for the regular synthetic dataset, but distributed even more heterogeneously, such that each client only had data from 1, 4, 7, 10 or 16 clusters, respectively. See fig. \ref{fig:ci_dist}.
\\\\
Figure \ref{fig:ci_results} shows that our method attains a similar average performance as compared to the central case, however with a larger variation. This is probably caused by differences in initializations. If the algorithm initializes in such a way that clients assign data to more clusters than what is being present in their data, the algorithm has a hard time correcting for that. Furthermore, it does not help that one client only has ten datapoints in total, meaning it initializes ten clusters of size one, of which none are being send to the central server due to privacy issues. Regardless, our method does outperform the algorithm from Dennis et. al. . This is likely due to our algorithm's ability to change the value of k for its local k-means step between clients.
\\\\



\begin{figure}[!htb]
\centering
    \subfigure[]{
        \includegraphics[width=0.47\textwidth]{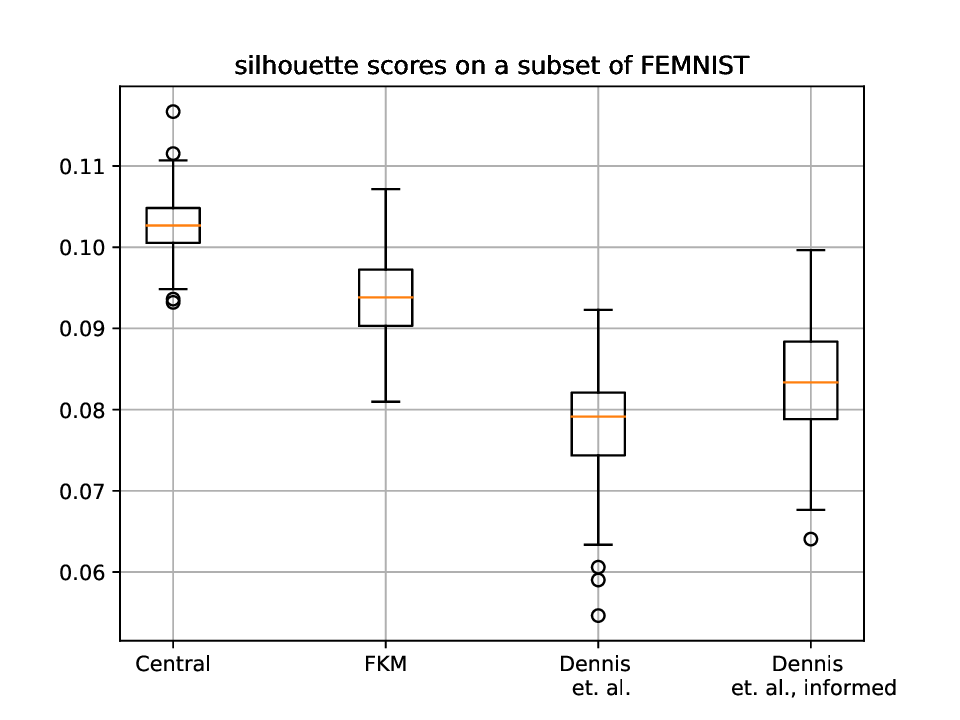}
        \label{fig:FEMNIST_silh}
     }
     \subfigure[]{
        \includegraphics[width = 0.47\textwidth]{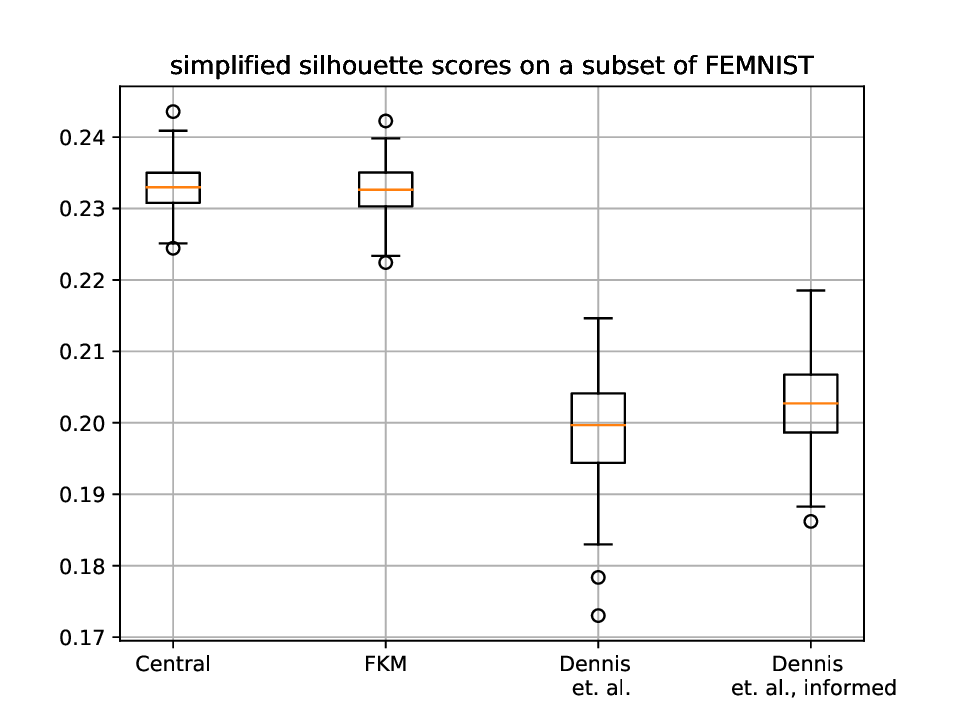}
        \label{fig:FEMNIST_ssilh}
     }
     \caption{Results on (a subset of) FEMNIST. (a) shows the silhoutte score, (b) the simplified silhouette score.}
\end{figure}

\noindent \textbf{Clustering higher dimensional real data.} \\
So far, all our experiments have been done on two dimensional, simulated data. For many use cases, however, data has a much higher dimensionality. In order to determine performance on a higher dimensional dataset, the Federated Extended MNIST (FEMNIST) from LEAF (\cite{LEAF}) (having a dimensionality of 784) was used, which separates the original Extended MNIST (\cite{EMNIST}) handwritten numbers and letters based on the person who wrote them. FEMNIST has a dimensionality of 784. This leaves approximately 110 datapoints per client; see appendix \ref{sec:FEMNIST_dist} for the distribution. Only 10 clients were used from the original FEMNIST, as this drastically sped up the experiments, while keeping enough data for a meaningful assessment. We set k = 60, in line with earlier experiments from Dennis et. al. Figure \ref{fig:FEMNIST_silh} shows that our method outperforms both settings of the method from Dennis et. al. . There is still a difference with a central clustering, however. This could be due to the relatively small amount of samples per client compared to the amount of dimensions, decreasing the quality of the local clusters.
\\\\
The FEMNIST experiments use the silhouette score as their performance metric. The silhouette score involves calculating distances from each point in a dataset to each other point in a dataset. This means that, to calculate a 'global' silhouette score, distances between datapoints from different clients need to be determined, something that can not be done in a straightforward federated manner. In our case, the simulated federated environment made it possible to calculate the silhouette score for evaluation purposes. In a real-life setting, the simplified silhouette score (\cite{simpl_silh}) could be a suitable alternative, as it only calculates distances between datapoints and cluster means, something which can be done on all clients separately.
\\\\
We compare the simplified silhouette score with the silhouette score from the same experiments in Figure \ref{fig:FEMNIST_ssilh}. There seems to be a high correlation between the two scores for a given method, which is in line with previous work \cite{ssilh_analysis}.

\newpage
\section{Discussion and conclusion}



This work describes the implementation and validation of a federated k-means clustering algorithm (FKM), enabling clustering over multiple datasets without sharing the underlying data. Our results show performances close to a central method, in which all data is brought into a single location. There are still some scenarios in which our method shows larger variability in performance as compared to a central clustering, however. These are mostly the more difficult scenarios, such as when there is an extreme distribution in the amount of cluster present on each client, or when the data has a high dimensionality as with the FEMNIST experiment. Assessment of our method on more heterogeneous and 'real life' datasets is therefore an important direction for future work. Nevertheless, FKM has shown to be a promising method in finding similarities among distributed datasets without the need of sharing any data.




%
\printbibliography

\appendix

\section{Background on k-means and k-means++} \label{sec:background}

\subsection{k-means clustering}
The objective of a clustering algorithm is to partition a given dataset into several subsets with similar features. The k-means clustering algorithm does so by trying to minimize the within cluster sum-of-squares criterion:
\begin{equation} \label{eq:inert}
    F_{km} = \sum_{j = 0}^m \min_{C_i \in C}(||X_j - C_i||^2)
\end{equation}
with $m$ the amount of samples, $C_i$ the cluster mean of cluster $i$, $C$ the set of all cluster means and $X_{j,k}$ being data point $j$ assigned to cluster $k$. The procedure in which the k-means algorithm tries to minimize equation \ref{eq:inert} consists of two steps. First, all data points get assigned to the cluster mean according to the lowest euclidean distance. Then, the mean center point from all points assigned to a certain cluster is calculated. This is done for every cluster, creating a new set of means to start the next round with. This process is repeated until the change within these means is smaller than a certain threshold (and the algorithm has reached convergence) (\cite{KMeans}).
\subsection {k-means++}
One of the drawbacks of classical k-means clustering is that its initialization is sampled uniformly from the underlying data. This means that having initial cluster means that all come from the same cluster is as probable as having initial cluster means spread across all clusters. Although the K-means algorithm itself can somewhat compensate for this, it still leads to large variability in performance. Arthur and  Vassilvitskii developed an initialization method for K-Means to combat this high variability, called k-means++ \cite{KMeans++}. Instead of sampling K cluster means from the data with uniform probability, datapoints get weighted based on their distance to the closest already mean that is chosen, with larger distances giving larger weights. This results in (on average) initializations that are more distributed over the space, and prevents (on average) initial cluster means from starting very close to each other, decreasing k-means performance.

\newpage
\section{Extra results on increasing amount of points per cluster} \label{sec:exresults}

\begin{figure}[!htb]
    \centering
    \subfigure[]{
        \includegraphics[width=0.3\textwidth]{figures/50ppc_b01_noises.eps}
        
        }
    \subfigure[]{
        \includegraphics[width=0.3\textwidth]{figures/50ppc_b1_noises.eps}
        
    }
    \subfigure[]{
        \includegraphics[width=0.3\textwidth]{figures/50ppc_b10_noises.eps}
        
    }
    \subfigure[]{
        \includegraphics[width = 0.3\textwidth]{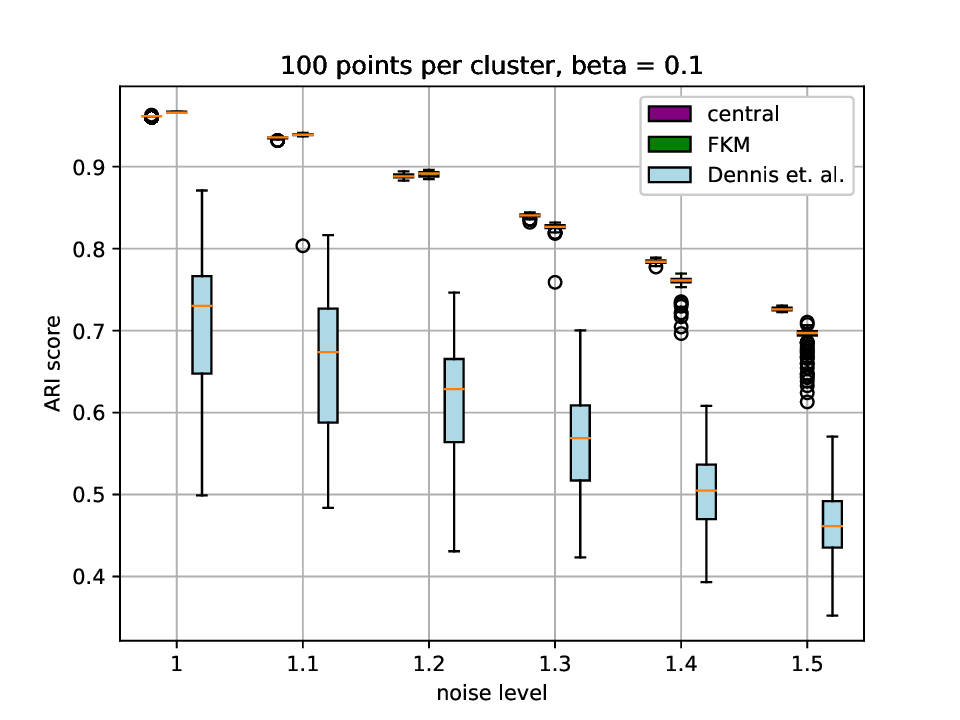}
    }
    \subfigure[]{
        \includegraphics[width = 0.3\textwidth]{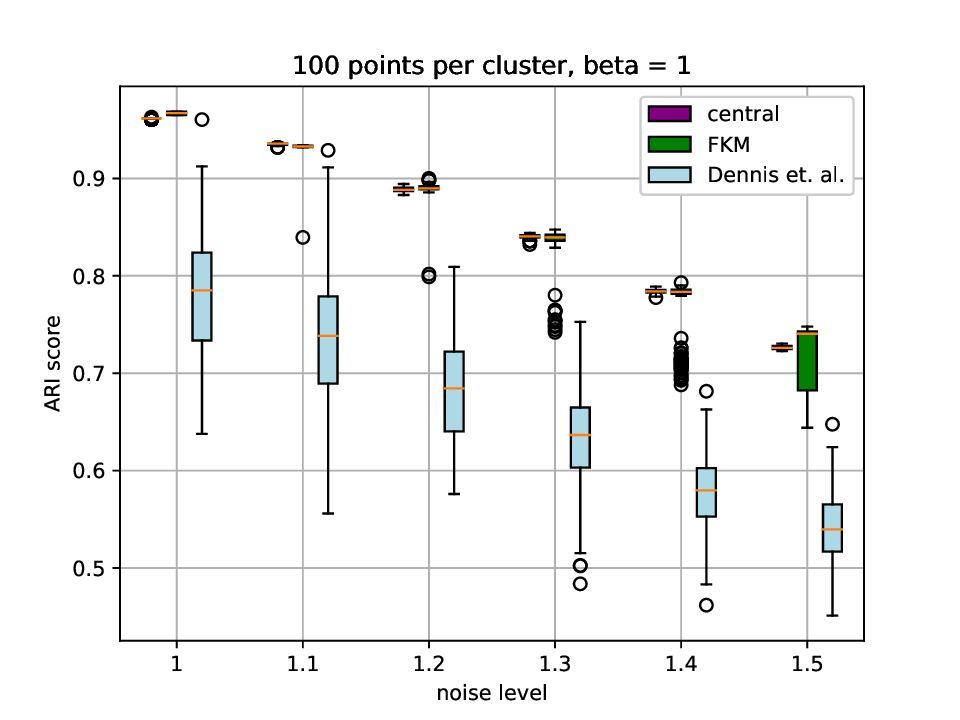}
    }
    \subfigure[]{
        \includegraphics[width = 0.3\textwidth]{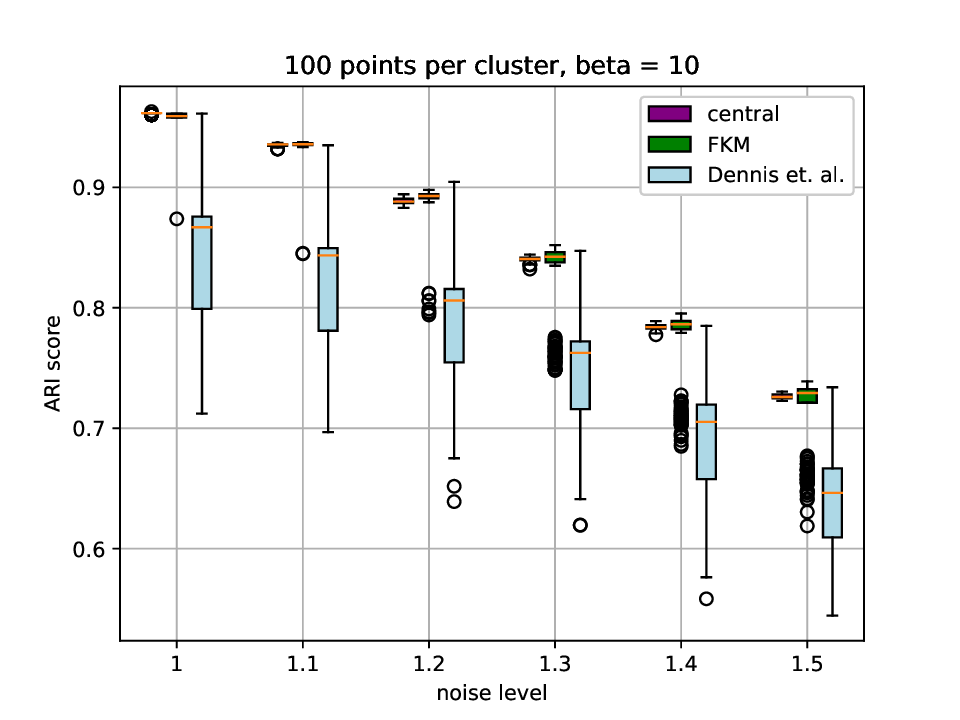}
    }    
    \subfigure[]{
        \includegraphics[width=0.3\textwidth]{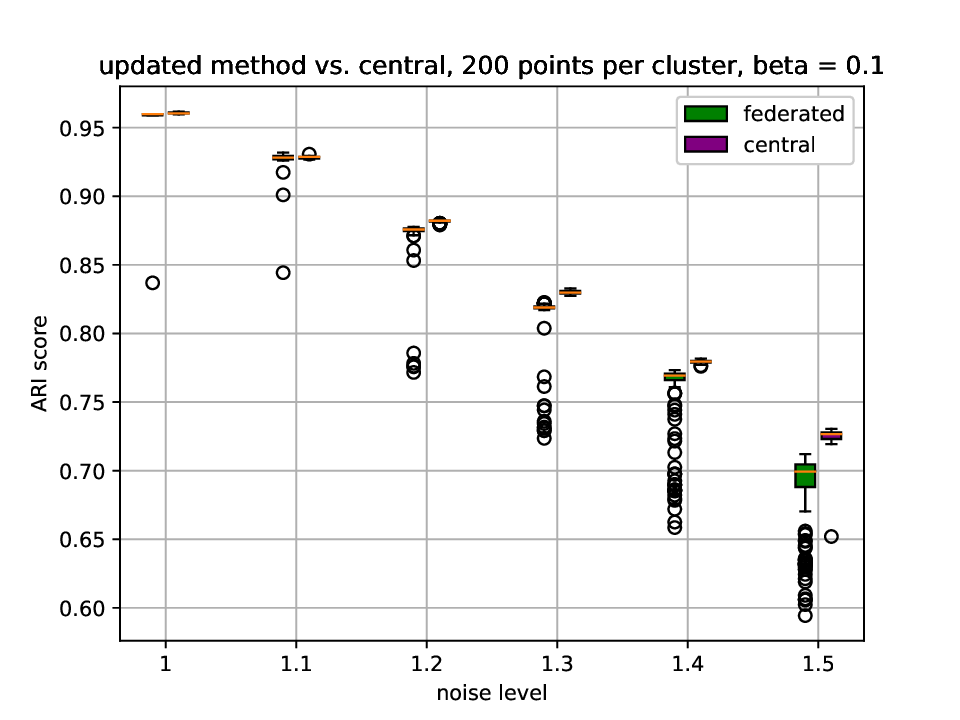}
        
        }
    \subfigure[]{
        \includegraphics[width=0.3\textwidth]{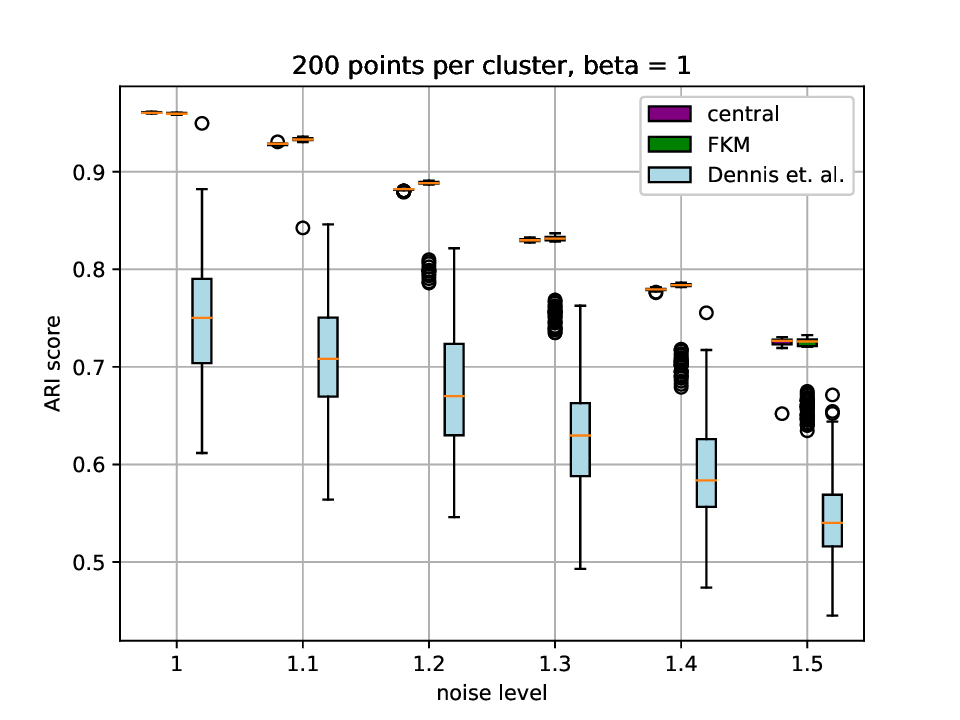}
        
    }
    \subfigure[]{
        \includegraphics[width=0.3\textwidth]{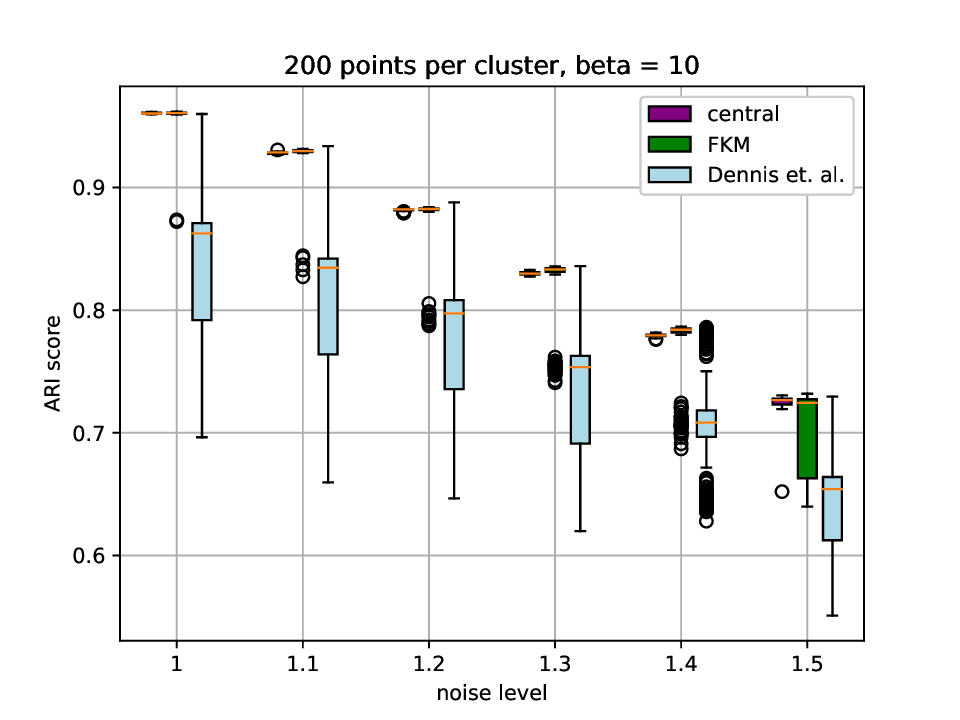}
        
    }
    \caption{results on using different levels of noise for different values of $\beta$ , with differing amounts of points per cluster. fom left to right, the columns correspond to $\beta$ = 0.1, 1 and 10 respectively. From top to bottom, the rows correspond to 50, 100, and 200 points per cluster.}\label{fig:abl_app}
\end{figure}

\newpage

\section{FEMNIST distribution} \label{sec:FEMNIST_dist}
\begin{figure}[!htb]
    \centering
    \includegraphics[width = 0.5\textwidth]{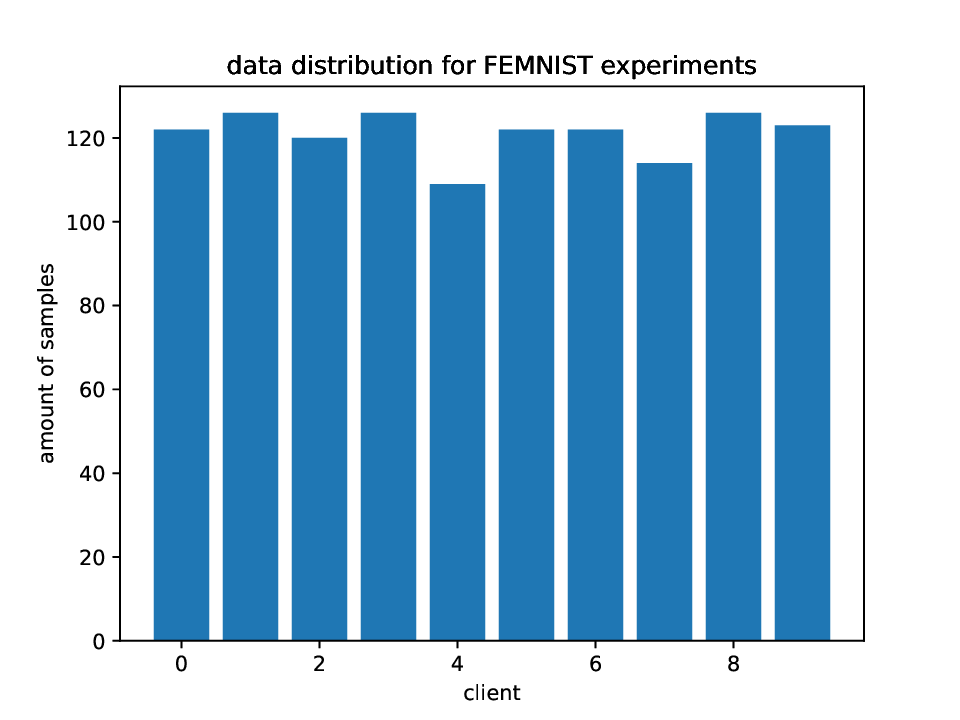}
    \caption{Sample distribution for the FEMNIST dataset}
    \label{fig:FEMNIST_dist}
\end{figure}

\end{document}